\documentclass[aip,cha,sd,amsmath,amsthm,amssymb,reprint]{revtex4-1}
\usepackage{graphicx,epsfig,pstricks,pst-node}
\usepackage{lipsum}
\usepackage{subcaption}
\usepackage{setspace,ulem}
\usepackage{times}
\usepackage{bm}
\usepackage{comment}
\usepackage{svg}
\usepackage{lineno}

\renewcommand{\thefootnote}{\fnsymbol{footnote}}

\usepackage{kotex}

\begin{document}

\title{Signal-noise separation using unsupervised reservoir computing}

\author{Jaesung Choi}
{\email{joseph9463@kias.re.kr}}
\affiliation{Center for Artificial Intelligence and Natural Sciences,
	Korea Institute for Advanced Study,
	Seoul 02455, South Korea}

\author{Pilwon Kim*}
{\email{pwkim@unist.ac.kr}}
\affiliation{Department of Mathematical Sciences, 
	Ulsan National Institute of Science and Technology, 
	Ulsan 44919, South Korea}

\date{\today}

\begin{abstract}
Removing noise from a signal without knowing the characteristics of the noise is a challenging task. This paper introduces a signal-noise separation method based on time series prediction. We use Reservoir Computing (RC) to extract the maximum portion of ``predictable information" from a given signal. Reproducing the deterministic component of the signal using RC, we estimate the noise distribution from the difference between the original signal and reconstructed one. The method is based on a machine learning approach and requires no prior knowledge of either the deterministic signal or the noise distribution. It provides a way to identify additivity/multiplicativity of noise and to estimate the signal-to-noise ratio (SNR) indirectly. The method works successfully for combinations of various signal and noise, including chaotic signal and highly oscillating sinusoidal signal which are corrupted by non-Gaussian additive/multiplicative noise. The separation performances are robust and notably outstanding for signals with strong noise, even for those with negative SNR.
\end{abstract}

\keywords{denoising, data filtering, time series data, reservoir computing, echo state network}

\maketitle

\section{INTRODUCTION}

Recovering the signal using data contaminated by noise is a long-standing challenge in signal and image processing \cite{G1, G2, G3, G4}. Estimation of the noise distribution is a key factor in determining the effectiveness of noise removal algorithms. In general, such algorithms are designed to work well for a specific distribution of noise. For example, algorithms that are designed to remove Gaussian noise will not perform as well for non-Gaussian noise. When handling denoising without prior knowledge of the noise distribution, it is often necessary to try a variety of algorithms such as low-pass filtering, wavelet-based smoothing and nonlinear adaptive filtering, and see which one works best \cite{G5, G6}. However, those trials may not be able to remove the noise effectively and often introduce artifacts to the recovered signal.

The signal-noise separation task becomes even more challenging if there is no reliable noise model. As the methods for dealing with multiplicative noise significantly differ from those employed for additive noise. Identifying whether the signal is corrupted by additive noise, multiplicative noise, or a combination of both is crucial for accurate signal analysis \cite{GP1, GP2}.

In this paper we consider the separation of signal from data contaminated by noise with unknown structure. More precisely, no assumption is made on the noise model or the noise distribution, and the separation is solved based only on the corrupted signal observations. The method proposed in this paper is based on time series prediction. Due to recent advances in machine learning, various time series analysis techniques have been proposed, including methods based on neural networks, and it is now possible to find deterministic patterns effectively even in a single time series. This allows us to redefine the signal filtering problem as a signal reconstruction problem.

Treating a given signal as training data, we instruct the machine predictor to discover as much predictable patterns within the signal as possible. If the predictor is not biased and its capacity is properly optimized within a range not to overfit data, the predictor can extract the deterministic portion of the signal successfully. By comparing the signal reconstructed through this process to the original one, we can identify whether the noise is additive or multiplicative, and can effectively approximate the target signal and the noise distribution together.

The proposed method was tested on several types of signal and noise. We focused on the separation of seemingly noise-like target signals under various non-Gaussian noises: this includes discrete/continuous chaotic signal with additive multi-modal noise and highly oscillating signal with multiplicative gamma noise.

The paper is organized as follows. In the next section we formulate the separation problem. Section \ref{sep_sec} explains the core procedures of the signal-noise separation. We deal with a reservoir computing method in Section \ref{rc_sec} which serves as a time series predictor in this paper. In Section \ref{ide_sec}, by applying the preceding procedures, we identify whether the noise is additive or multiplicative, and then obtain its distribution accordingly. Section \ref{val_sec} deals with  validation to optimize the hyperparameters of the predictor.  In Section \ref{exp_sec}, the proposed method is applied to combinations of various signal and noise. The results are compared to those from conventional filters.

\section{Problem formulation}\label{for_sec}

\begin{figure*}[!htb]
        \includegraphics[width=1\linewidth]{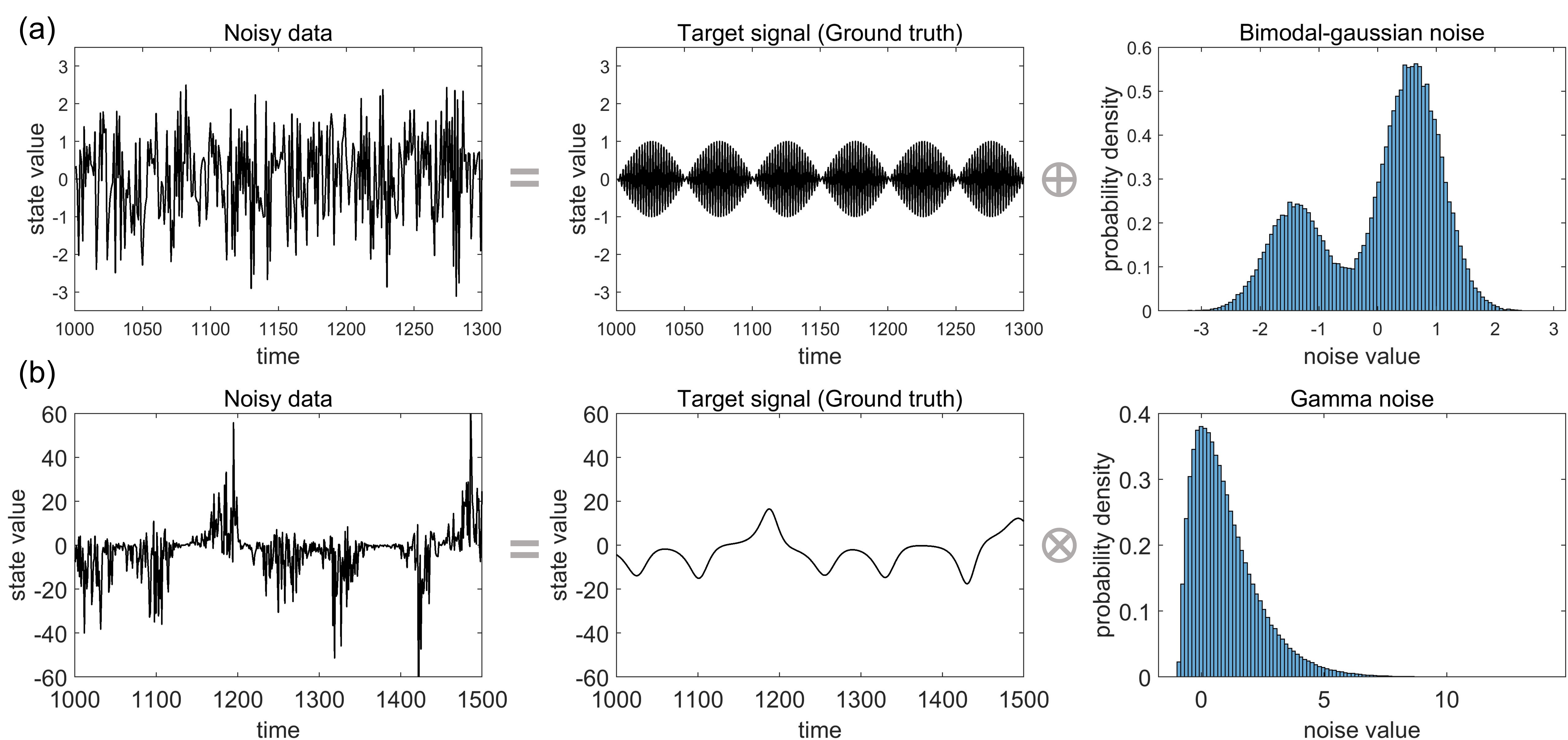}
	\caption{singal-noise separation for (a) additive noise and (b) multiplicative noise}
        \label{fig:figure1}
\end{figure*}

Let \( x_i; i=0,1, \ldots, N \) be the signal of interest for noise separation. We assume the signal follows either the observational model for additive noise
\begin{equation}
x_i = q_i + \xi_i, \quad E[\xi_i]=0,
\end{equation}
or the model for multiplicative noise
\begin{equation}
x_i = q_i \xi_i, \quad E[\xi_i]=1.
\end{equation}
Here \( q_i \) is the deterministic signal component and \( \xi_i \) is the noise component at the \( i \)th sampling instant. We assume no prior knowledge on the deterministic mechanism for \( q_i \) nor on the noise distribution for \( \xi_i \).

By comparing the given signal \( x_i \) with the signal \( q_i \) reconstructed from a machine learning method, we characterize the form of noise \( \xi_i \) either as (1) or (2), and then estimate its distribution accordingly. 

Figure 1 illustrates two examples of separation of the signal: (a) highly oscillating sinusoidal signal with additive bimodal-gaussian noise and (b) chaotic signal with multiplicative gamma noise. This separation procedure will be handled in detail in the following sections.

\section{Outline of the separation procedure} \label{sep_sec}

We treat the given signal \( x_i, 0 \leq i \leq N \) to be processed as a dataset for learning, and split it into a training part \( x_i, 0 \leq i \leq K \) and a validation part \( x_i, K < i \leq N \). We adopt a \( m \)-step time series predictor \( P \) based on a machine learning method which supposedly reproduces one step ahead from previous steps as \( x_i = P(x_{i-1}, x_{i-2}, \dots, x_{i-m}), m \leq i \leq N \) after learning the signal.

The signal-noise separation for \( x_i, m \leq i \leq N \) with respect to \( P \) proceeds in the following six steps.

\begin{enumerate}
    \item Training the predictor \( P \) to minimize the cost function over the training part,
    \begin{align*}
        \sum_{i=m}^{K} \left\| x_i - P(x_{i-1}, x_{i-2}, \ldots, x_{i-m}) \right\|^2.
    \end{align*}

    \item Reconstruction of \( q_i \) using the predictor \( P \):
    \begin{align*}
        \hat{q}_i = P(x_{i-1}, x_{i-2}, \ldots, x_{i-m}), \quad m \leq i \leq N.
    \end{align*}

    \item Evaluation of misfits :
    \begin{align*}
        \psi_i = x_i - P(x_{i-1}, x_{i-2}, \ldots, x_{i-m}) = x_i - \hat{q}_i, \quad m \leq i \leq K.
    \end{align*}

    \item Identification of additivity/multiplicativity of noise from the correlation between \( \psi_i \) and \( \hat{q}_i \) 

    \item Approximation of the noise distribution \( D \):
    \begin{align*}
        \hat{\xi}_i &= \psi_i , \quad \text{ for } m \leq i \leq K. \\
        (\text{or } \hat{\xi}_i &= \psi_i / \hat{q}_i, \text{ in case of multiplicative noise})
    \end{align*}

    \item Evaluation of the approximate error over the validation part:
    \begin{align*}
        \sum_{i=m}^{K}  \left\| x_i - \hat{q}_i \right\|^2. 
    \end{align*}

\end{enumerate}

Note that the time series predictor \( P \) is not actually used to predict \( x_i, i > N \), but rather to reconstruct the original time series \( x_i, m \leq i \leq N \). Considering the deterministic nature of pattern prediction, we expect that the training process ``neutralizes'' the effect of noise \( \xi_i \) and yields estimation of the deterministic portion \( q_i \).

Step 6 is necessary to optimize the hyperparameters of the predictor \( P \). We repeat Step 1 through 6 using a hyperparameter-tuning method such as Bayesian optimization, until the validation error reaches its minimum. The validation process will be discussed in more detail in Section \ref{val_sec}.

\section{Reservoir Computing Predictor}\label{rc_sec}

The predictor $P$ in the process outlined in the previous section can be implemented with 
any common machine learning methods. 
Reservoir computing (RC) is one of the popular choices to deal with time series due to its simple architecture and dynamic nature \cite{RC1}. 
RC comprises two main components: (i) a ``reservoir" is a fixed nonlinear recurrent network  and (ii) a ``readout" is a trainable linear output layer.  

We especially use a simple discrete type of RC, Echo State Networks (ESN) to predict the time series $x_i, i=0,1,\cdots$. ESN consists of $L$ nodes whose temporal states \( \mathbf{r}(i) \in \mathbb{R}^L\) evolves through the equation:
\begin{equation}\label{esn}
    \mathbf{r}(i+1) = (1 - \alpha)\mathbf{r}(i) + \alpha \tanh(\mathbf{A}\mathbf{r}(i) + \mathbf{W}_{\text{in}}x_{i-1}), i=1,2,\cdots,
\end{equation}
where $\alpha$ is a leaking rate, \( \mathbf{A} \in \mathbb{R}^{L \times L} \) the internal weight matrix,  and  \( \mathbf{W}_{\text{in}} \in \mathbb{R}^{L} \) the input weight vector.
For more general introduction to ESN, the readers are referred to \cite{RC2, RC3}. 
The readout weight vector \( \mathbf{W}_{\text{out}} \) is determined in the training process by solving:
\begin{equation}
    \mathbf{W}_{\text{out}} = \arg\min_{\mathbf{W} \in \mathbb{R}^N} \left( \sum_{i=1}^{K} \left\| x_{i+1} - \mathbf{W}^T\mathbf{r}(i)\right\|^2 + \lambda \|\mathbf{W}\|_F^2 \right),
\end{equation}
where \( \lambda \) is a regularization parameter.

Once \( \mathbf{W}_{\text{out}} \) is obtained, we can estimate the deterministic signal as
\begin{equation}
    \hat{q}_i = \mathbf{W}_{\text{out}}\mathbf{r}(i), i=0,1, \cdots, N.
\end{equation}
to implement the separation process described in Section 3. We call this Signal-Separation using Reservoir Computing (SSRC).


\section{Identification of additivity /multiplicativity of noise}\label{ide_sec}

\begin{figure}[!htb]
	\includegraphics[width=1\linewidth]{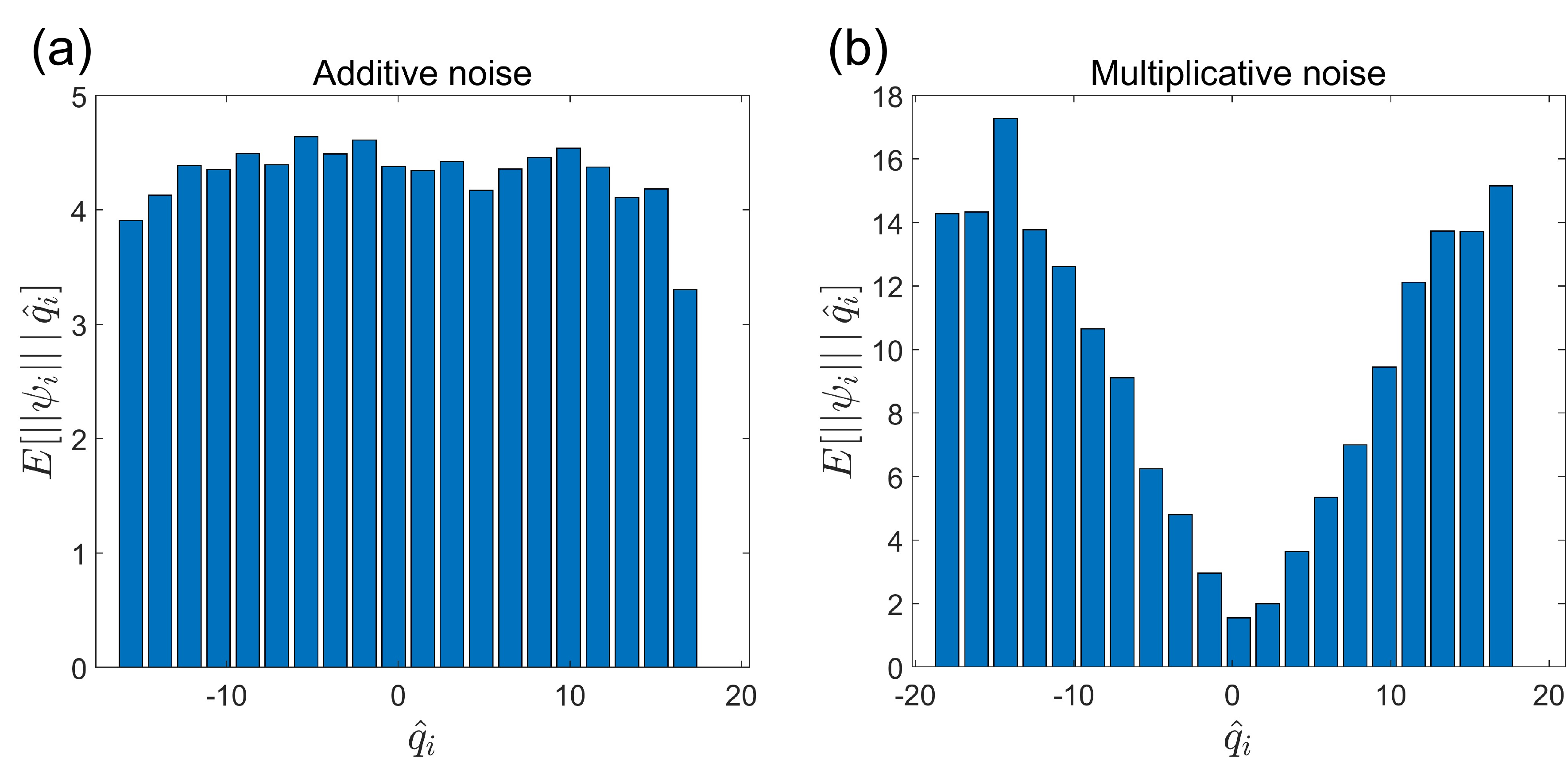}
	\caption{Graphs of \( E[\left\| \psi_i \right\| \mid \hat{q}_i] \) with respect to \( \hat{q}_i \): (a) additive noise (b) multiplicative noise.}
	\label{fig:figure9}
\end{figure}

We can discriminate between additive and multiplicative noise, by studying the conditional distribution of the misfit \( \psi_i \), given \( \hat{q}_i \). More specifically, the type of noise can be inferred from the graph of \(E[\left\| \psi_i \right\| | \hat{q}_i]\) with respect to \(\hat{q}_i\). We use the approximation \(E[\left\| \psi_i \right\| | \hat{q}_i] = E[\left\| x_i - \hat{q}_i \right\| | \hat{q}_i] \approx E[\left\| x_i - q_i \right\| | \hat{q}_i]\). In case of additive noise as in Eq. (1), we have
\[ E[\left\| x_i - q_i \right\| \mid q_i] = E[\left\| \xi_i \right\| \mid q_i] = E[\left\| \xi_i \right\|] = \textsl{const.}, \]
since noise \( \xi_i \) is independent from the deterministic signal \( q_i \). This leads to a flat graph as in Figure 2(a). On the contrary, multiplicative noise \( \xi_i \) in Eq. (2) gives
\begin{align*}
E[\left\| x_i - q_i \right\| \mid q_i] &= E[\left\| q_i \right\| \left\| \xi_i - 1 \right\| \mid q_i] \\
&= \left\| q_i \right\| E[\left\| \xi_i - 1 \right\|] \approx C\left\| q_i \right\|,
\end{align*}
where \( C = E[\left\| \xi_i - 1 \right\|] = \textsl{const.} \) Figure 2(b) shows an example of the corresponding symmetric V-shaped graph with a vertex at the origin.

Once confirmed as additive noise, the approximate noise \( \hat{\xi}_i \) and its distribution can be obtained by \( \hat{\xi}_i = \psi_i \). In case of multiplicative noise, we set \( \hat{\xi}_i = x_i / \hat{q}_i \).

\section{Validation to avoid overfitting}\label{val_sec}

\begin{figure}[!htb]
	\includegraphics[width=1\linewidth]{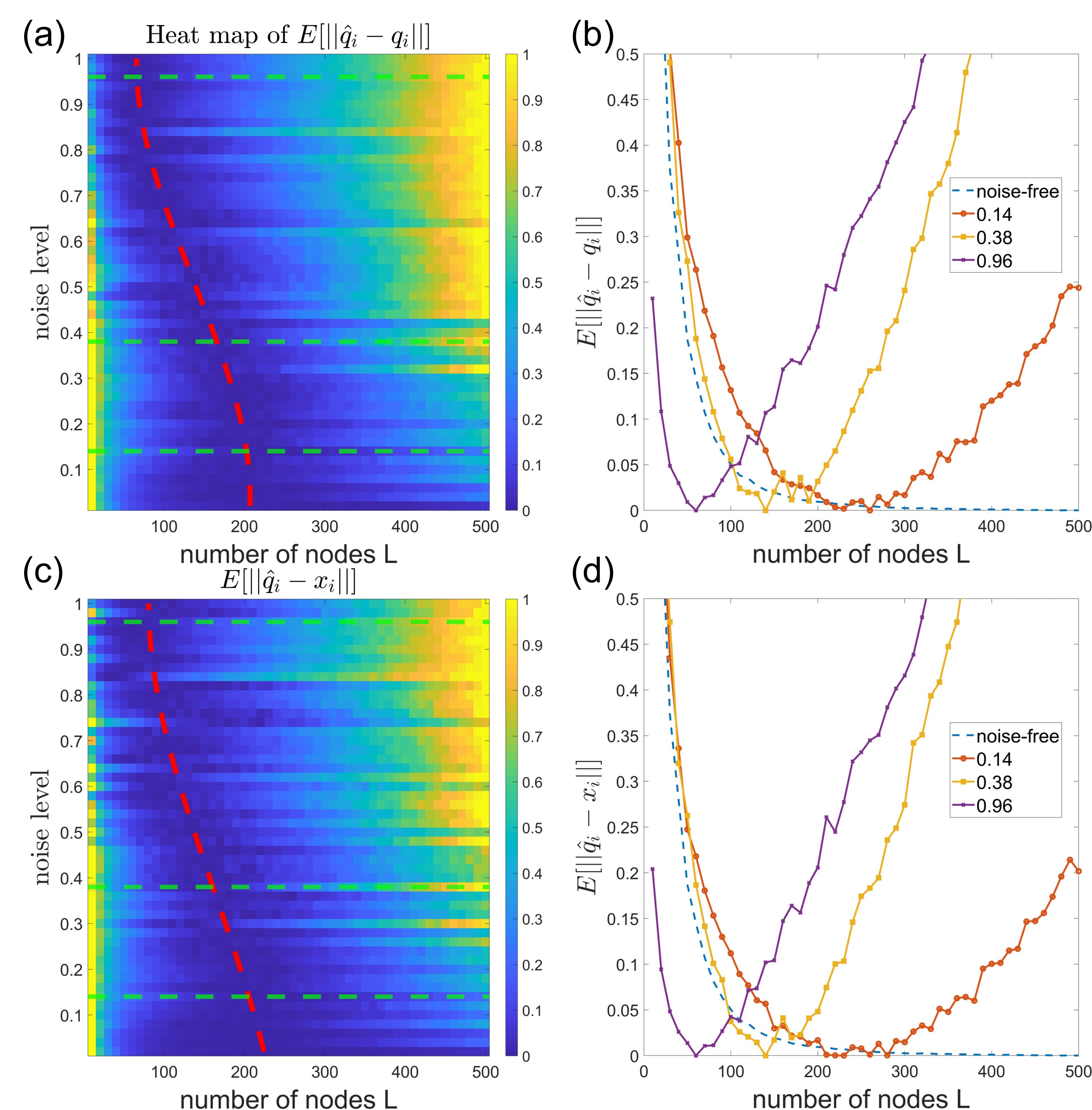}
	\caption{(Color online) Optimal reservoir size according to the noise level. (a) Heat map of  $ E[\left\| \hat{q}_i - \ q_i \right\|] $ with respect to the reservoir size and the noise level (b) Horizontal cross sectional graphs of (a) at the noise level 0.14, 0.38 and 0.96 (13.92, 5.28 and -2.9dB in SNR, respectively). (c) Heat map of $E[\left\| \hat{q}_i - \ x_i \right\| ]$ with respect to the reservoir size and the noise level (d) Horizontal cross sectional graphs of (c) at the noise level 0.14, 0.38 and 0.96. The red dotted lines in (a) and (c) indicate the optimal node numbers at each noise level. For convenience of comparison, the heat maps are normalized in a way that the minimum on each horizontal cross sectional line is placed at zero. }
	\label{fig:figure3}
\end{figure}

The goal of training a predictor \( P \) for a given signal \( x_i \) is to set \( P \) to discover the maximum deterministic patterns, that is, \( q_i \). In Section \ref{exp_sec}, we will see through a number of examples that the reservoir-based predictor of reasonable size can successfully extract \( q_i \), separating it from various noise. However, \( P \) is commonly exposed to risk of overfitting, since it uses only the single signal for training, the very signal that is the subject of the separation. If \( P \) has excessive computational capacity, it is likely to overfit in estimating \( q_i \),  futilely looking for spurious patterns in the noise.

Such overfitting may arise especially when the deterministic portion of the signal is small compared to noise, that is, when the signal-to-noise ratio (SNR) is small. In other words, at low SNRs, the use of overly powerful predictors can lead to overfitting, which degrades the reliability of the reconstructed signal.

Figure 3(a) is a heat map of the approximation error $\left\|\hat{q}_i-q_i \right\| $ showing this trend. The x-axis represents the number of nodes used for RC, and the y-axis corresponds to the noise level of $\xi_i$, which linearly determines the standard deviation of $\xi_i$.

One can determine the optimal number of nodes at a noise level, by comparing the error values along the corresponding horizontal line across the heat map. In Figure 3(b), such three exemplary cross sectional graphs are plotted at the noise level 0.14, 0.38 and 0.9, respectively. All the cross sectional graphs are vertically shifted for the minimum to be placed at zero for convenience of comparison. This shows the optimal choice for the node numbers are 260, 140 and 70, respectively.

The red dotted line in Figure 3(a) indicates where the optimal choice for the reservoir size (node numbers) occurs at each level of the noise. Since the number nodes used in the reservoir roughly indicates its computational capacity, the optimal node number tends to decrease as the noise level increases. 

Note that the such optimal choice in Figure 3(a) and (b) was actually determined by comparing the approximate deterministic component \( \hat{q}_i \) to the true one \( q_i \). However, \( q_i \) is not available in practice. Thus we use the error between the given signal \( x_i \) and the reproduced signal \( \hat{q}_i \) over the validation set instead. Figure 3(c) and (d) shows that the validation error between \( \hat{q}_i \) and \( x_i \) largely follows the tendency of the errors between \( \hat{q}_i \) and \( q_i \), and therefore is capable to provide the criterion to tune the hyperparameters of the predictor. Refer to the supplement for more detail of the other parameter values used for the figures.

\section{Experiments}\label{exp_sec}
This section applies the separation procedure described in Section \ref{sep_sec} to various type of signals. We are especially interested in chaotic and highly oscillating target signal \( q_i \), which is often hard to identify from noise. More specifically, three deterministic signals are sampled from (a) Lorenz systems (b) high-frequency sinusoidal signal and (c) logistic map with memory (mLogistic). Refer to Supplement for more details of the signals including the generating equations.

We test combinations of these signals with the three types of non-Gaussian noise: (1) additive lognormal noise, (2) additive bimodal noise, and (3) multiplicative gamma noise. In all examples, the time series consists of 9000 data points. We allocate 1000 points of them to the validation set and use Bayesian search to determine hyperparameters such as the number of nodes $L$ and the spectral radius of the connection matrix $\mathbf{A}$.

We compare the separation results with those of conventional methods such as low-pass filters, wavelet filters, median filter and nonlinear adaptive filter. In nonlinear adaptive filtering method outlined by Gao et al. \cite{NA}, time series is segmented into subintervals and adaptively merged after polynomial fitting for enhanced signal clarity. For these filters, the noise distributions are obtained in a similar manner to the proposed method, that is, from the observation on difference between the original signal and the filtered signal.  More detailed information about the filters used for comparison can be found in Supplement. The root-mean square error (RMSE) is used to evaluate the reconstruction error for \( q_i \). To measure how close the distributions of \( \hat{\xi}_i \) is to that of \( \xi_i \), we use the Jensen-Shannon divergence (JSD).

\subsection{Signals with additive lognormal noise}
In the first set of examples, we test SSRC for signals corrupted by additive lognormal noise. 
Figure 4 shows a one of the separation results for the signal $x_i$ sampled from the Lorenz system. It is corrupted by additive noise in a one-sided lognormal distribution with SNR is 2.67dB. In Figure 4(a), the reconstructed signal \( \hat{q}_i \) by SSRC is plotted. Figure 4(b) is the separation results of the low pass filter 
whose cut off frequency is the lowest 25\% of the spectrum, and 4(c) is those of the wavelet filter (Daubechies-4).
It is confirmed that SSRC performs better than the other two methods both in reconstruction of the deterministic and noise component.

\begin{figure}[!htb]
    \includegraphics[width=\linewidth]{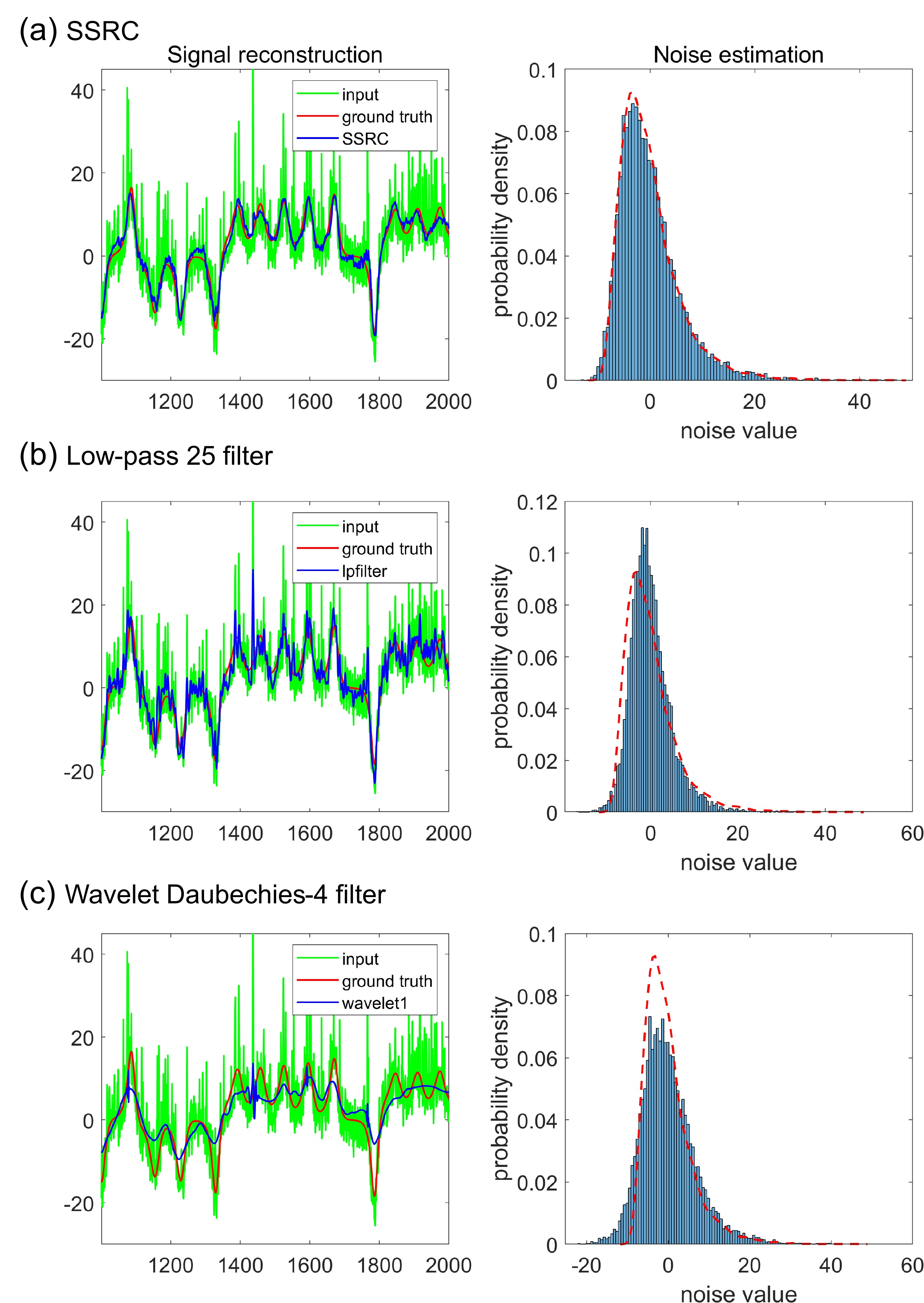}
    \caption{(Color online) Lorenz signal \( x_i \) corrupted by additive lognormal noise with SNR=2.67(dB), (a) separation results of SSRC, (b) low pass filtering, (c) wavelet based filtering}
    \label{fig:figure4}
\end{figure}

Table 1 contains thorough results from the various combinations for the lognormal noise. The second and third rows show the separation performances for the highly oscillating signal and the logistic map with memory, respectively, both of which are corrupted by lognormal noise. In all three cases, we can see that SSRC outperforms the other filter methods in reconstruction of the deterministic signal.

\begin{table}[ht]
\centering
\begin{tabular}{lccc}
\hline
 & Lorenz & High frequency & mLogistic \\
\hline
SSRC & \textbf{0.1786} & \textbf{0.0452} & \textbf{0.0661}\\
Wavelet1 & 0.4865 & 1.0001 & 0.3789\\
Wavelet2 & 0.4871 & 1.0001 & 0.3750\\
Low-pass Filtering25 & 0.3349 & 1.0031 & 0.2738\\
Low-pass Filtering50 & 0.4786 & 1.0063 & 0.1712\\
Low-pass Filtering75 & 0.6039 & 1.0099 & 0.1278\\
Median Filtering & 0.3624 & 0.2216 & 0.2659\\
Adaptive filtering & 0.1994 & 1.0007 & 0.3784\\

\hline
\end{tabular}
\caption{Mean reconstruction errors for additive log-normal noise with various data types of average SNR values of 2.67 dB, 15.3 dB, and 21.6 dB for Lorenz, high-frequency, and mLogistic system, respectively, out of 100 realizations.}

\label{table:comparison}
\end{table}

\begin{figure}[!htb]
    \includegraphics[width=0.9\linewidth]{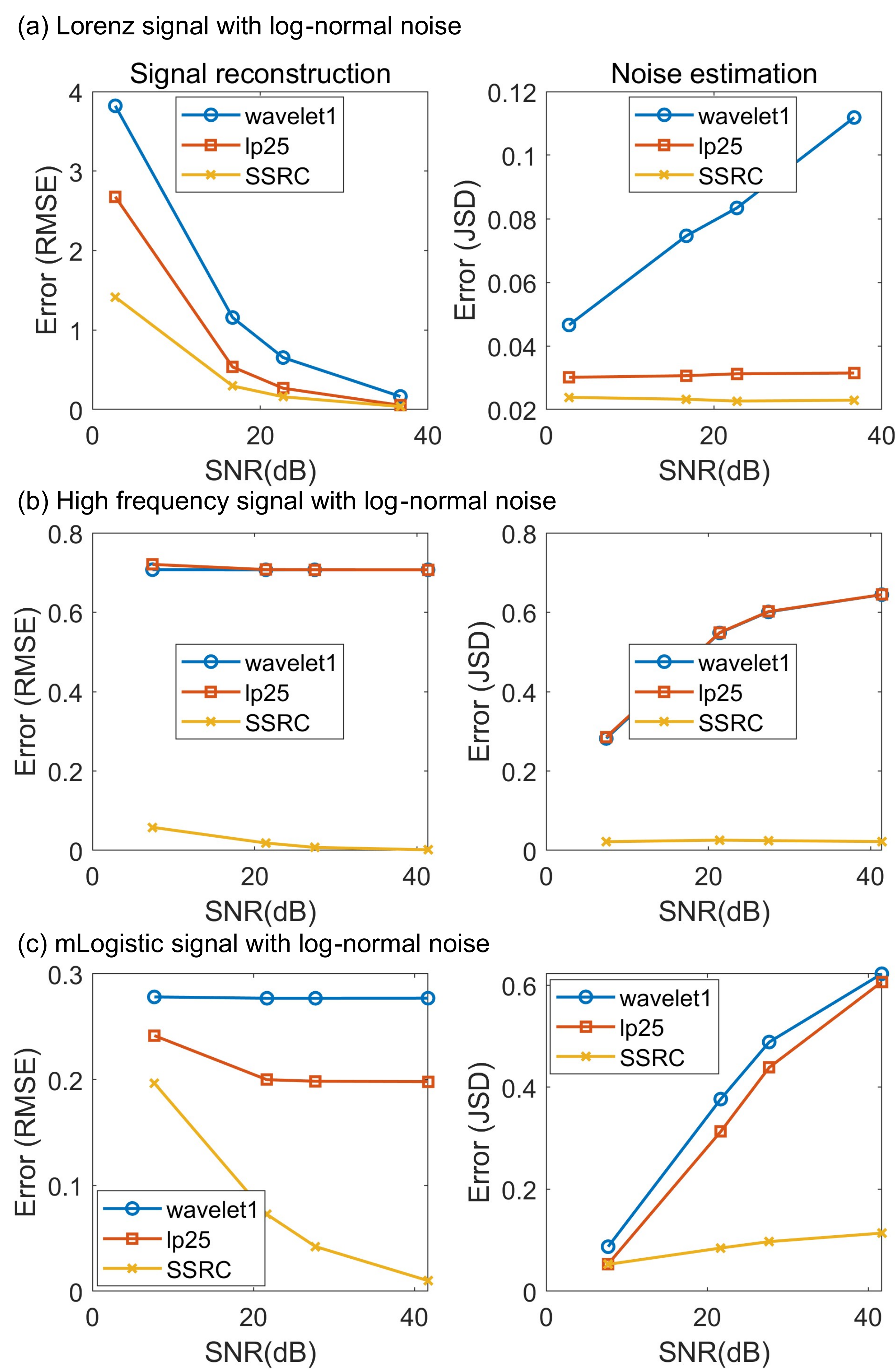}
    \caption{ Comparison of separation performance for additive log-normal noise according to SNR. (a) Lorenz signal (b) Highly oscillating sinusoidal signals}
    \label{fig:figure5}
\end{figure}

Figure 5 compares the separation performances  according to the noise level. Figure 5(a), (b) and (c) shows errors in separation of three types of signals, respectively, corrupted by additive lognormal noise with different levels of SNR. In all bins of the comparison, SSRC outperforms the other filters. We can see that the accuracy in recovering the deterministic component increases as SNR increases. It is notable that the performance of SSRC is almost insensitive to SNR, i.e., it is less affected by the noise level in separation of signals. This forms a striking difference from other filters which have substantial difficulty distinguishing between signal and noise, especially when SNR is small.

\subsection{Signals with additive mutimodal noise}

Multimodal noise often arises when measurement systems integrate input from several sources. In the second set of examples, we test SSRC for signals corrupted by multimodal noise. Fig 6 shows one of the separation results. The original signal is a highly oscillating sinusoidal signals corrupted by additive bimodal noise with 4.58(dB). It is clear that the reconstruction by SSRC in Figure 6(a) excels those by the low pass filter and the wavelet filter in Figures 6(b) and 6(c), respectively. Neither of two filters can reproduce a simple deterministic sinusoidal pattern with high frequency. In addition, SSRC successfully captures bimodality of the noise distribution, while the others fail to detect it.

Table 2 also confirms the separation performance of SSRC is better than the other conventional filters in three types of signals corrupted by the bimodal noise. Figure 7 compares the performance of methods according to SNR. Once again, SSRC turned out to be better than the other filters in all bins of the comparison.

\begin{figure}[!htb]
	\includegraphics[width=1\linewidth]{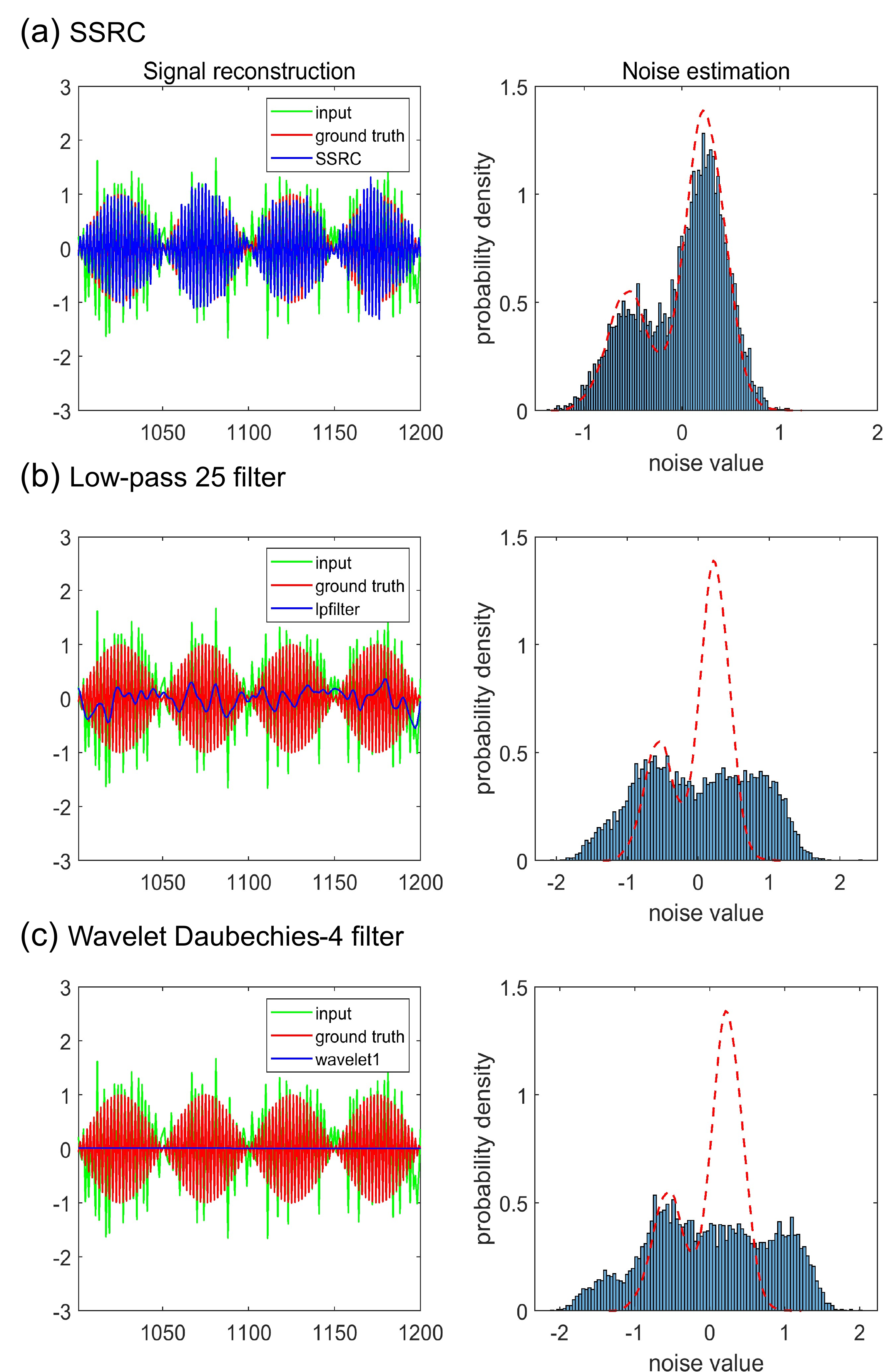}
	\caption{(Color online) Highly oscillating sinusoidal signals \( x_i \) corrupted by additive bimodal-gaussian noise with SNR=4.58(dB), (a) separation results of SSRC, (b) low pass filtering, (c) wavelet based filtering}
        \label{fig:figure6}
\end{figure}

\begin{table}[ht]
\centering
\begin{tabular}{lccc}
\hline
 & Lorenz & High frequency & mLogistic \\
\hline
SSRC & \textbf{0.4811} & \textbf{0.1578} & \textbf{0.3136}\\
Wavelet1 & 0.8749 & 1.0008 & 0.3809\\
Wavelet2 & 0.8708 & 1.0006 & 0.3804\\
Low-pass Filtering25 & 1.1563 & 1.0359 & 0.3768\\
Low-pass Filtering50 & 1.6529 & 1.0722 & 0.4075\\
Low-pass Filtering75 & 2.0871 & 1.1127 & 0.4843\\
Median Filtering & 1.5841 & 0.6020 & 0.4497\\
Adaptive filtering & 0.5221 & 1.0070 & 0.3952\\

\hline
\end{tabular}
\caption{Mean reconstruction errors for additive bimodal-gaussian noise with various data types of SNR values of -8.01 dB, 4.58 dB, and 5.1 dB for Lorenz, high-frequency, and mLogistic system, respectively, out of 100 realizations.}
\label{table:comparison}
\end{table}

\begin{figure}[!htb]
	\includegraphics[width=0.9\linewidth]{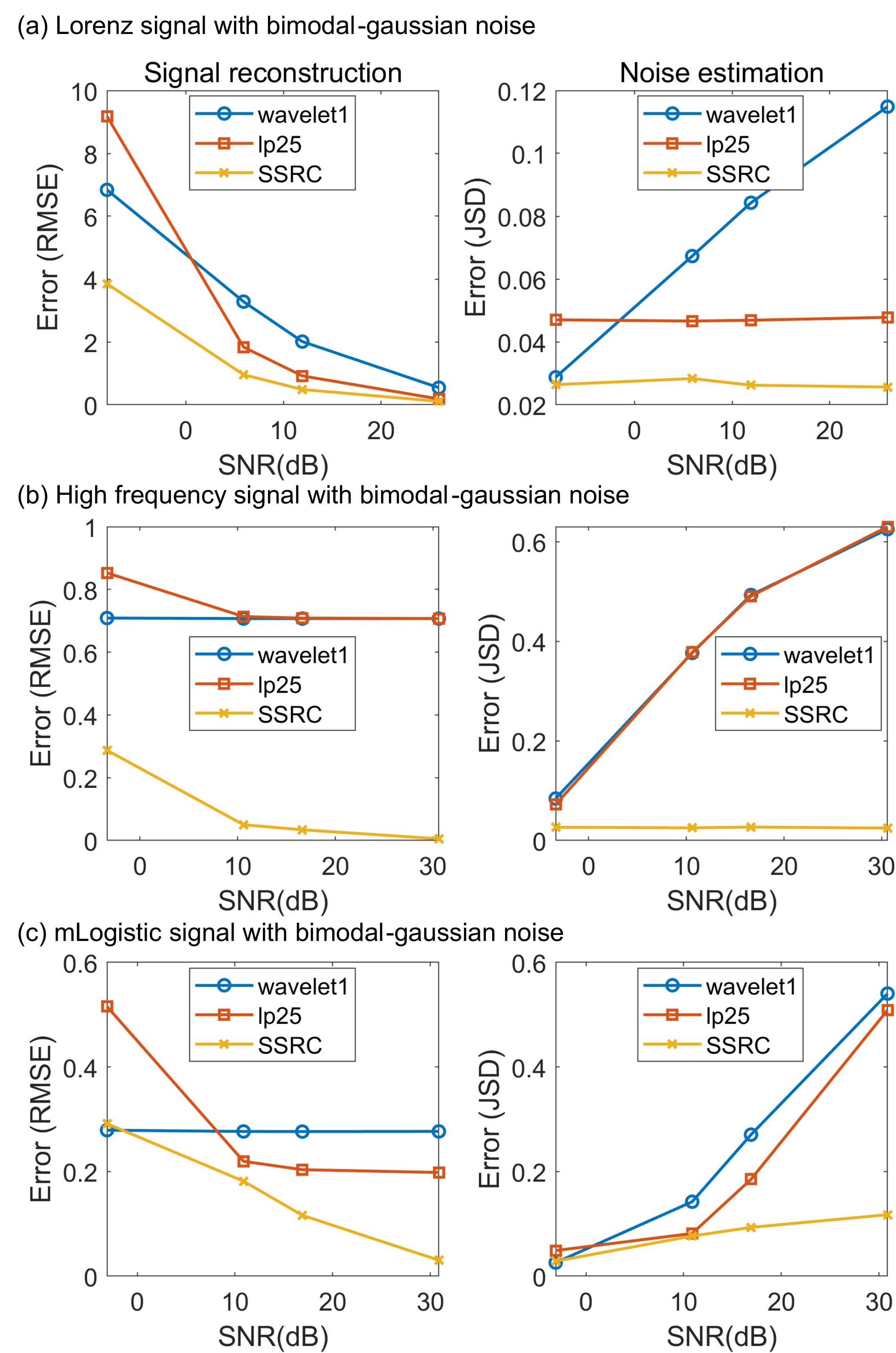}
	\caption{ Comparison of separation performance for bimodal noise according to SNR. (a) Lorenz signal corrupted by additive bimodal noise (b) Highly oscillating signal corrupted by additive bimodal noise}
	\label{fig:figure7}
\end{figure}

\subsection{Signals with multiplicative gamma noise}
Multiplicative noise has been observed in many signal processing applications, biological movement systems and optical systems \cite{M1,M2,M3,M4}. Removal of multiplicative noise is challenging because it distorts the amplitude of a signal substantially. In the third set of examples, we apply SSRC to signals corrupted by multiplicative noise. Figure 8 shows one of the separation results for a signal from the logistic map with memory. The signal is corrupted by the multiplicative noise with 9(dB) that follows a gamma distribution with mean 1. Figure 8(a), (b) and (c) compare the reconstructed signals \( \hat{q}_i \) and the distribution of the error \( \hat{\xi}_i \) by SSRC, the low pass filter and the wavelet filter, respectively. Note that the results of SSRC in the plots (a) accompany with verification of the multiplicativity of the noise through the graph in Figure 2(b). On the contrary, one can find no clue about the noise type in processing the other filters. It is clear again that SSRC outperforms two other filters in separating \( q_i \) and \( \xi_i \). Refer to Table 3 and Figure 9 for further comparison for other signal-noise combinations at various noise levels. 

\begin{figure}[!htb]
	\includegraphics[width=1\linewidth]{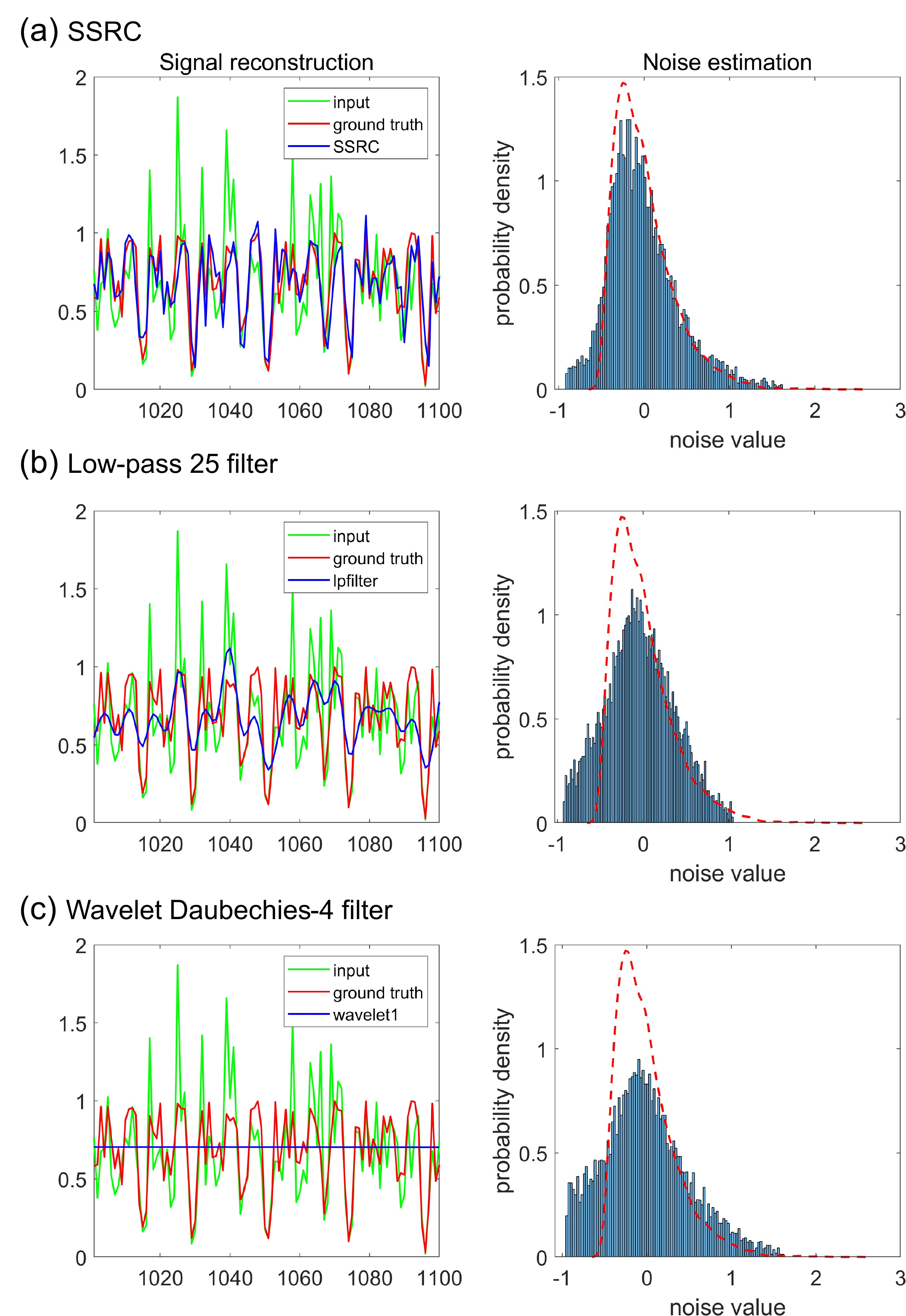}
 	\caption{(Color online) mLogistic signals \( x_i \) corrupted by multiplicative gamma noise with SNR=9(dB), (a) separation results of SSRC, (b) low pass filtering, (c) wavelet based filtering}
	\label{fig:figure10}
\end{figure}

\begin{table}[ht]
\centering
\begin{tabular}{lccc}
\hline
 & Lorenz & High frequency & mLogistic \\
\hline
SSRC & \textbf{0.2867} & \textbf{0.0670} & \textbf{0.1920}\\
Wavelet1 & 0.6371 & 1.0001 & 0.3788\\
Wavelet2 & 0.6337 & 1.0001 & 0.3780\\
Low-pass Filtering25 & 0.6205 & 1.0129 & 0.3159\\
Low-pass Filtering50 & 0.8883 & 1.0263 & 0.2828\\
Low-pass Filtering75 & 1.1218 & 1.0415 & 0.3118\\
Median Filtering & 0.7114 & 0.3224 & 0.3339\\
Adaptive filtering & 0.3031 & 1.0026 & 0.3848\\

\hline
\end{tabular}
\caption{Mean reconstruction errors for multiplicative gamma noise with various data types of SNR values of -2.68 dB, 9 dB, and 9 dB for Lorenz, high-frequency, and mLogistic system, respectively, out of 100 realizations.}
\label{table:comparison}
\end{table}

\begin{figure}[!htb]
	\includegraphics[width=0.9\linewidth]{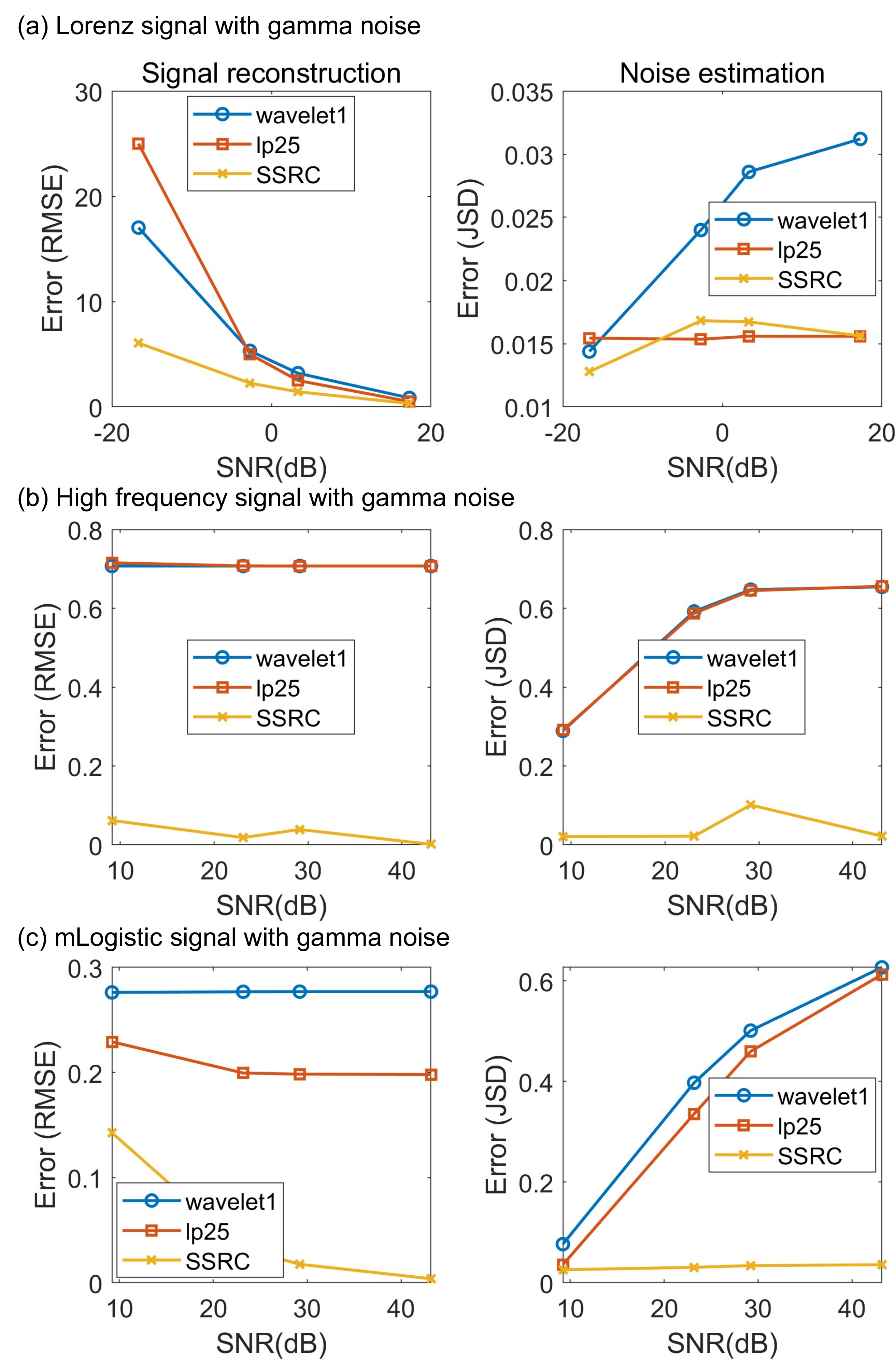}
	\caption{ Comparison of separation performance for multiplicative gamma noise according to SNR. (a) Lorenz signal (b) highly oscillating sinusoidal signals.  In the right panels of each figure, the multiplicative noise distribution is estimated by the equation $\hat{\xi}_i=\psi_i/q_i$, and compared to the true one in JSD. In such estimation, $1\%$ of $q_i$ with the smallest magnitude is discarded to avoid potential overflow.}
	\label{fig:figure9}
\end{figure}



\section{Discussion}
The method operates without prior knowledge of noise characteristics and effectively isolates deterministic signals from noise with diverse non-Gaussian distributions, encompassing symmetric and multimodal forms. Its efficacy has been validated across a range of signal and noise combinations, including chaotic signals contaminated with multiplicative gamma noise and highly oscillatory sinusoidal signals corrupted by intense additive noise.

The proposed method is flexible and simple, and at its core it exploits the ability of machine learning to maximize the extraction of deterministic patterns. The effect of noise is neutralized in the process of finding patterns in the signal, and then its distribution is recovered by comparing the restored pattern back to the original time series. The method can be considered as unsupervised learning in the sense that the predictor requires only single time series, the one that is subject to filtering. We used RC in this paper as a predictor, but there is no intrinsic difference in using any other popular predictors such as multi-layer perceptron or LSTM.

To avoid overfitting, we left a portion of the given signal as a validation set and used it to optimize the predictor's performance. The capacity of the predictor determined in this way allows us to indirectly estimate SNR of the signal. In other words, the optimal capacity of the predictor, like the node number of the reservoir, can be regarded as an indirect measure of the amount of deterministic information in the signal. 

This paper mainly focused on separation of signals with additive and multiplicative noise. However, in practice, the signals are often corrupted by combination of those two types of error together. Separation of such noise requires more refined applications of time series prediction and will be studied in the future work.

\bigskip

\noindent{\bf Methods}\\
All computations were carried out using MATLAB R2023a version.
The simulation of the data system was conducted using the RK4 (Runge-Kutta 4th order method).
Low-pass filtering, wavelet-based filtering, and Bayesian optimization were all implemented using MATLAB's built-in functions.
Detailed explanations about parameter settings and the system equation can be found in the supplementary material. \\

\noindent{\bf Data availability}\\
The authors declare that the data supporting the findings of this study can be recreated as described in the manuscript and also obtainable from the corresponding author upon request.\\

\noindent{\bf Acknowledgements}\\
\\

\noindent{\bf Author contributions}\\
J.C. and P.K. conceived of the presented idea. J.C. developed the theory and performed the computations. P.K. supervised the project and wrote the manuscript.k\\

\noindent{\bf Competing interests}\\
Authors declare no competing interests.\\

\noindent {\bf Correspondence} and requests for materials should be addressed to Pilwon Kim.\\

\noindent{\bf References}
\bibliographystyle{IEEEtranN}
\bibliography{ref}

\end{document}


\maketitle

\section{Approximation in Validation Process}

In case of additive noise:
\begin{align*}
E[(x_i-\hat{q}_i)^2] &= E[(q_i+\xi_i - \hat{q}_i)^2]\\
&= E[(q_i-\hat{q}_i)^2]+2E[q_i-\hat{q}_i]E[\xi_i]+E[\xi_i^2] \\
&= E[(q_i-\hat{q}_i)^2]+Var[\xi_i].
\end{align*}

Hence the minimum of $E[(x_i-\hat{q}_i)^2]$ coincides with that of $E[(q_i-\hat{q}_i)^2]$ with respect to $\hat{q}_i$.
\\

In case of multiplicative noise:

\begin{align*}
E[(x_i-\hat{q}_i)^2] &= E[(q_i\xi_i - \hat{q}_i)^2]\\
&= E[q_i^2]E[\xi_i^2]-2E[q_i\hat{q}_i]E[\xi_i]+E[\hat{q}_i^2] \\
&= E[q_i^2](Var[\xi]+1)-2E[q_i\hat{q}_i]+E[\hat{q}_i^2]\\
&= E[q_i^2]Var[\xi]+E[(q_i-\hat{q}_i)^2],
\end{align*}

which implies $E[(x_i-\hat{q}_i)^2]\approx E[(q_i-\hat{q}_i)^2]$ if we assume $Var[\xi]\ll 1.$

\section{Parameter Setting}

In our study, we compared SSRC's performance with various noise filtering methods each of which is optimized for the corresponding task. For implementation of those filters, we primarily use MATLAB's built-in functions except for the nonlinear adaptive filtering. The detailed configurations for SSRC and comparative methods are as follows:

\begin{enumerate}
    \item[(a)] \textbf{SSRC:} MATLAB's \texttt{bayesopt} function is used to optimize  six hyperparameters, including $pr$ and $\rho$. Here, $pr$ is the sparsity level of the reservoir's internal connection matrix $A$ and $\rho$ is its spectral radius. These parameters are crucial for both the dynamics and the memory capacity of the reservoir. 

    \begin{table}[!htbp]
\centering
\caption{Hyperparameter settings for SSRC}
\label{tab:hyperparameters}
\begin{tabular}{@{}lp{6cm}l@{}}
\toprule
\textbf{Hyperparameter} & \textbf{Description}                                  & \textbf{Range}            \\
\midrule
\(\alpha\)              & Leaking rate                                          & 0.01 to 1                 \\
\(L\)                   & Number of nodes in the reservoir                      & 10 to 1000                \\
\(\lambda\)             & Regularization strength                               & \(10^{-10}\) to 1, log scale \\
$pr$                    & Sparsity level of matrix \(A\)                        & 0.05 to 1                 \\
$\sigma_{in}$           & Scaling factor for \(W_{in}\)                         & 0.1 to 2.0                \\
\(\rho\)                & Spectral radius of matrix \(A\)                       & 0.3 to 1.5                \\
\bottomrule

\end{tabular}
\end{table}

    \item[(b)] \textbf{Wavelet based Filtering:} We utilized MATLAB's \texttt{wdenoise} function for Daubechies 4 (\texttt{db4}) and Symlets 4 (\texttt{sym4}) wavelet filters. The function applies  the universal thresholding and soft threshold rules to balance noise suppression and signal fidelity.

    \item[(c)] \textbf{Low-pass Filtering:} We applied low-pass filters using MATLAB's \texttt{butter} and \texttt{filtfilt} functions, with the cutoff frequencies set at 25\%, 50\%, and 75\% of the Nyquist frequency, respectively. The sampling rate was assumed to be 100 Hz.
    

    \item[(d)] \textbf{Median Filtering:} MATLAB's \texttt{medfilt1} function was used with a window size of 5.


\item[(e)] \textbf{Adaptive Filtering:} To effectively employ the adaptive denoising algorithm, it is essential to determine the polynomial order \(K\) and the segment length \(\omega\). Following the criteria established in previous studies \cite{NA, NA2}, we predominantly opted for \(K=3\) and \(\omega=33\). However, there are several cases where the expected well-shaped graphs delineated in the literature do not appear, due to different noise distribution or nature of the data. In such cases, we try to maintain the specified parameters of \(K=3\) and \(\omega=33\) to ensure methodological consistency across the analysis.

\end{enumerate}

\section{Data Generation}

For our analysis, we generated datasets using different dynamical systems known for their chaotic behavior and complex dynamics. The following subsections describe the systems and the parameters used to generate the respective datasets.

\subsection{Lorenz System}
The Lorenz system is a three-dimensional dynamical system that exhibits chaotic flow, characterized by the famous Lorenz attractor. The equations are as follows:
\begin{align*}
\frac{dx}{dt} &= \sigma(y-x), \\
\frac{dy}{dt} &= x(\rho - z) - y, \\
\frac{dz}{dt} &= xy - \beta z.
\end{align*}
The standard parameter values used are $\sigma = 10$, $\rho = 28$, and $\beta = \frac{8}{3}$.

\subsection{Highly oscillating sinusoidal signal}
To simulate a signal with mixed frequency content, we combined low and high-frequency sine waves as follows:
\begin{align*}
x(t) = \sin(2\pi f_1 t) + 0.5 \sin(2\pi f_2 t).
\end{align*}
The low frequency $f_1$ was set to 5 Hz, and the high frequency $f_2$ was set to 1000 Hz.

\subsection{Logistic Map with Memory}
The Logistic map with memory is a generalization of the classic Logistic map defined as:
\begin{align*}
x_{n+1} = r \tilde{x}_n (1 - \tilde{x}_n),
\end{align*}
where $\tilde{x}_n=c_0 x_n+ c_1 x_{n-1} + \cdots + c_l x_{n-l}$ is the weighted sum of the previous values. Here the weights $c_0, c_1, \cdots, c_l$ are non-negative and their summation is 1. In our case, we used $\tilde{x}_n=c_1 x_{n-1}+ c_2 x_{n-2} + c_3 x_{n-3}$ where the coefficients are randomly chosen as $[c_1, c_2, c_3] = [0.5577, 0.4133, 0.0290]$.
\\

Each of these systems has been chosen for their unique characteristics, which are well-suited for testing the robustness of signal-noise separation techniques in unsupervised reservoir computing.